\documentclass[11pt,a4paper]{article}
\PassOptionsToPackage{breaklinks}{hyperref}
\usepackage[hyperref]{acl2019}
\usepackage[utf8]{inputenc}
\usepackage{times}
\usepackage{latexsym}
\usepackage{color, colortbl}

\usepackage{graphicx}
\usepackage{multirow}
\usepackage{tabularx}
\usepackage{breakurl}

\newenvironment{citemize}{\begin{list}{$\bullet$}{\topsep=.5\smallskipamount\itemsep=0pt\parsep=1pt\labelwidth=.5em}}{\end{list}}

\aclfinalcopy
\def\aclpaperid{308} 

\title{Neural Architectures for Nested NER through Linearization}

\author{Jana Strakov\'{a} \and Milan Straka \and Jan Haji\v{c}\\
  Charles University \\
  Faculty of Mathematics and Physics \\
  Institute of Formal and Applied Linguistics \\
  {\tt \{strakova,straka,hajic\}@ufal.mff.cuni.cz} \\}

\date{}

\begin{document}
\maketitle

\begin{abstract}
  We propose two neural network architectures for nested named entity recognition~(NER), a setting in which named entities may overlap and also be labeled with more than one label. We encode the nested labels using a linearized scheme. In our first proposed approach, the nested labels are modeled as multilabels corresponding to the Cartesian product of the nested labels in a standard LSTM-CRF architecture. In the second one, the nested NER is viewed as a sequence-to-sequence problem, in which the input sequence consists of the tokens and output sequence of the labels, using hard attention on the word whose label is being predicted. The proposed methods outperform the nested NER state of the art on four corpora: ACE-2004, ACE-2005, GENIA and Czech CNEC.  We also enrich our architectures with the recently published contextual embeddings: ELMo, BERT and Flair, reaching further improvements for the four nested entity corpora. In addition, we report flat NER state-of-the-art results for CoNLL-2002 Dutch and Spanish and for CoNLL-2003 English.
\end{abstract}

\section{Introduction}
\label{section:introduction}

In nested named entity recognition, entities can be overlapping and
labeled with more than one label such as in the example \textit{``The Florida
Supreme Court''} containing two overlapping named entities \textit{``The
Florida Supreme Court''} and \textit{``Florida''}.\footnote{Example from
ACE-2004 \cite{Doddington}, \url{https://catalog.ldc.upenn.edu/LDC2005T09}.}

Recent publications on nested named entity recognition involve stacked LSTM-CRF
NE recognizer \cite{Ju2018}, or a construction of a special structure that
explicitly captures the nested entities, such as a constituency graph
\cite{Finkel2009} or various modifications of a directed hypergraph
\cite{Lu2015,Katiyar2018,Wang2018}.

We propose two completely neural network architectures for nested nested named
entity recognition which do not explicitly build or model any structure and
infer the relationships between nested NEs implicitly:

\begin{citemize}
  \item In the first model, we concatenate the nested entity multiple labels
    into one multilabel, which is then predicted with a standard LSTM-CRF
    \cite{Lample2016} model. The advantages of this model are simplicity and
    effectiveness, because an already existing NE pipeline can be reused to
    model the nested entities. The obvious disadvantage is a large growth of NE
    classes.
  \item In the second model, the nested entities are encoded in a sequence and
    then the task can be viewed as a sequence-to-sequence (seq2seq) task, in
    which the input sequence are the  tokens (forms) and the output sequence
    are the labels. The decoder predicts labels for each token, until a special
    label \texttt{"<eow>"} (end of word) is predicted and the decoder moves to
    the next token.
\end{citemize}

The expressiveness of the models depends on a non-ambiguous encoding of the
nested entity structure. We use an enhanced BILOU scheme described in
Section~\ref{section:linearization}.

The proposed models surpass the current nested NER state of the art on four
nested entity corpora: ACE-2004, ACE-2005, GENIA and Czech CNEC. When the
recently introduced contextual embeddings -- ELMo \cite{Peters2018}, BERT
\cite{BERT} and Flair \cite{Akbik} -- are added to the architecture, we reach
further improvements for the above mentioned nested entity corpora and also
exceed current state of the art for CoNLL-2002 Dutch and Spanish and for
CoNLL-2003 English.


\section{Related Work}
\label{section:related_work}

\citet{Finkel2009} explicitly model the nested structure as a syntactic
constituency tree.

\citet{Ju2018} run a stacked LSTM-CRF NE recognizer as long as at least one
nested entity is predicted, from innermost to outermost entities.



\citet{Wang2018} build a hypergraph to capture all possible entity mentions in
a sentence.

\citet{Katiyar2018} model nested entities as a directed hypergraph similar to
\citet{Lu2015}, using RNNs to model the edge probabilities.

Our proposed architectures are different from these works because they do not
explicitly build any structure to model the nested entities. The nested entity
structure is instead encoded as a sequence of labels, and the artificial neural
network is supposed to model the structural relationships between the named
entities implicitly.

A sequence-to-sequence architecture similar to one of our approaches is used by
\cite{liu-zhang-2017-shift} to predict the hierarchy of constituents in order
to extract lookahead features for a shift-reduce constituency parser.

\section{Datasets}
\label{section:datasets}

We evaluate our results on four nested NE corpora:

\begin{citemize}
  \item English \textbf{ACE-2004},
    \cite{Doddington}\footnote{\url{https://catalog.ldc.upenn.edu/LDC2005T09}}.
    We reuse the train/dev/test split used by most previous authors
    \cite{Lu2015,Muis2017,Wang2018}.
  \item English \textbf{ACE-2005}\footnote{\url{https://catalog.ldc.upenn.edu/LDC2006T06}}.
    Again, we use the train/dev/test split by \citet{Lu2015,Muis2017,Wang2018}.
  \item English \textbf{GENIA} \cite{GENIA}. We use the $90\%/10\%$ train/test split used
    by previous authors \cite{Finkel2009,Lu2015,Muis2017,Wang2018}.
  \item Czech \textbf{CNEC} -- Czech Named Entity Corpus 1.0. As previous authors \cite{Strakova2016}, we predict the $42$ fine-grained \textit{NE types} and $4$ \textit{containers} from the first annotation round.
\end{citemize}

We evaluate flat NER on these four languages: \textbf{CoNLL-2003} English and German \cite{CoNLL2003} and \textbf{CoNLL-2002} Dutch and Spanish \cite{CoNLL2002}.

In all cases, we use the train portion of the data for training and the
development portion for hyperparameter tuning, and we report our final results
on models trained on concatenated train+dev portions and evaluated on the test
portion, following e.g. \cite{Ratinov2009,Lample2016}.

Our evaluation is a strict one: each entity mention is considered correct only when both the span and class are correct.

\section{Methods}
\label{section:methods}

\subsection{Nested NE BILOU Encoding}
\label{section:linearization}

Our goal is to encode the nested entity structure into a CoNLL-like,
per-token BILOU encoding,\footnote{\texttt{B-} (beginning), \texttt{I-}
(inside), \texttt{U-} (unit-length entity), \texttt{L-} (last) or \texttt{O}
(outside) labels \cite{Ratinov2009}.} as in the following example for sentence
\textit{``in the US Federal District Court of New Mexico .''}:

\begin{small}
\begin{verbatim}
in              O
the             B-ORG
US              I-ORG|U-GPE
Federal         I-ORG
District        I-ORG|U-GPE
Court           I-ORG
of              I-ORG
New             I-ORG|B-GPE
Mexico          L-ORG|L-GPE
.               O
\end{verbatim}
\end{small}

The mapping from tokens to multilabels is defined by the two following rules:
(1) entity mentions starting earlier have priority over entities starting later, and
(2) for mentions with the same beginning, longer entity mentions have priority over shorter ones. A multilabel for a word is then a concatenation of all intersecting entity
mentions, from the highest priority to the lowest.


Another, more formalized look at the BILOU encoding is that it is a BILOU
encoding of an unfolded directed hypergraph similar to \citet{Katiyar2018}, in
which the shared entity labels are not collapsed and the \texttt{O} is used
only for tokens outside any entity mention.

We use a trivial heuristic during decoding, matching labels of consecutive words
by order only. Therefore, an \verb|I-| or \verb|L-| label is merged with a preceding
\verb|B-| or \verb|I-| if they appear on the same position in neighboring multilabels
and have the same type.

\subsection{Neural Models for Nested NER}

Both our models are encoder-decoder architectures:

\textbf{LSTM-CRF:} The encoder is a bi-directional LSTM and the decoder is a CRF \cite{Lample2016},
modeling multilabels from Section~\ref{section:linearization}.

\textbf{Sequence-to-sequence (seq2seq):} The encoder is a bi-directional LSTM
and the decoder is a LSTM. The tokens are viewed as the input sequence, and the
encoded labels are predicted one by one by the decoder, until the decoder
outputs the \texttt{"<eow>"} (end of word) label and moves to the next token. We use
a hard attention on the word whose label(s) is being predicted, and predict labels for a word from highest to lowest priority as defined in Section~\ref{section:linearization}.

We train the network using the lazy variant of the Adam optimizer \cite{Adam}, which only updates accumulators for variables that appear in the current
batch,\footnote{\texttt{tf.contrib.opt.lazyadamoptimizer} from
\url{www.tensorflow.org}} with parameters $\beta_1=0.9$ and $\beta_2=0.98$. We
use mini-batches of size~$8$. As a regularization, we apply dropout with rate
$0.5$ and the word dropout replaces $20\%$ of words by the unknown token to
force the network to rely more on context. We did not perform any complex
hyperparameter search.

In our baseline versions, we use the following word- and character-level word
embeddings:

\begin{citemize}
  \item pretrained word embeddings: For English, we train our own word
    embeddings of dimension $300$ with \texttt{word2vec}\footnote{Skip-gram,
    for tokens with at least $10$ occurrences, window $=5$, dimension $=300$,
    negative sampling $=5$.} on the English Gigaword Fifth
    Edition.\footnote{\url{https://catalog.ldc.upenn.edu/LDC2011T07}} For other
    languages (German, Dutch, Spanish and Czech) we use the FastText word
    embeddings \cite{FastText}.\footnote{\url{https://fasttext.cc/docs/en/crawl-vectors.html}}
  \item end-to-end word embeddings: We embed the input forms and lemmas ($256$ dimensions) and POS tags (one-hot).
  \item character-level word embeddings: We use bidirectional GRUs
    \cite{Cho2014,Graves2005} of dimension $128$ in line with \citet{Ling2015}: we represent every
    Unicode character with a vector of dimension $128$, and concatenate
    GRU outputs for forward and reversed word characters.
\end{citemize}

We further add contextual word embeddings to our baselines:

\begin{citemize}
  \item \textbf{+ELMo} \cite{Peters2018}: pretrained contextual word embeddings
    of dimension $512$ for English.
  \item \textbf{+BERT} \cite{BERT}: pretrained contextual word embeddings of dimension
    $1024$ for English\footnote{BERT-Large Uncased from \url{https://github.com/google-research/bert}} and $768$ for other languages\footnote{BERT-Base Multilingual Uncased from \url{https://github.com/google-research/bert}}. For each token, we generate the contextual word embedding
    by averaging all BERT subword embeddings in the last four layers \cite{BERT} without finetuning.
  \item \textbf{+Flair} \cite{Akbik}: pretrained contextual word embeddings of dimension
    $4096$ for all languages except Spanish.\footnote{Not yet available in December
    2018.}
\end{citemize}

We use the implementation provided by \citet{Akbik} to generate the Flair and
ELMo word embeddings.\footnote{\url{https://github.com/zalandoresearch/flair}}

We do not use any hand-crafted classification features in any of our models.

\begin{table*}[!t]
  \begin{small}
  \begin{center}
  \catcode`@ = 13\def@{\bfseries}
  \catcode`! = 13\def!{\itshape}
  \begin{tabular}{l|c|c|c|c}
    model                       & ACE-2004 & ACE-2005 & GENIA & CNEC 1.0 \\
    \hline
    \cite{Finkel2009}**         & --       & --       & 70.3  & --      \\
    \cite{Lu2015}**             & 62.8     & 62.5     & 70.3  & --      \\
    \cite{Muis2017}**           & 64.5     & 63.1     & 70.8  & --  \\
    \cite{Katiyar2018}          & 72.70    & 70.5     & 73.6  & --  \\
    \cite{Ju2018}*              & --       & 72.2     & 74.7  & --  \\
    \cite{Wang2018}             & 75.1     & 74.5     & 75.1  & --  \\
    \cite{Strakova2016}         & --       & --       & --    & 81.20   \\
    \hline
    LSTM-CRF             & 72.26  & 71.62  & !76.23 & 80.28  \\
    LSTM-CRF+ELMo        & !78.72 & !78.36 & !75.94 & --     \\
    LSTM-CRF+BERT        & !81.48 & !79.95 & !77.80 & !85.67 \\
    LSTM-CRF+Flair       & !77.65 & !77.25 & !76.65 & !81.74 \\
    LSTM-CRF+BERT+ELMo   & !80.07 & !80.04 & !76.29 & --     \\
    LSTM-CRF+BERT+Flair  & !81.22 & !80.82 & !77.91 & !85.70 \\
    LSTM-CRF+ELMo+BERT+Flair & !80.19 & !79.85 & !76.56 & -- \\
    \hline
    \rowcolor{gray!30}
    seq2seq              & !77.08 & !75.36 & !76.44 & !82.96 \\
    seq2seq+ELMo         & !81.94 & !81.95 & !77.33 & --     \\
    seq2seq+BERT         & !84.33 & !83.42 & !78.20 & !86.73 \\
    seq2seq+Flair        & !81.38 & !79.83 & !76.63 & !83.55 \\
    seq2seq+BERT+ELMo    & !84.32 & !82.15 & !77.77 & --     \\
    seq2seq+BERT+Flair   & @84.40 & @84.33 & @78.31 & @86.88 \\
    seq2seq+ELMo+BERT+Flair & !84.07 & !83.41 & !78.01 & --  \\
    \end{tabular}
    \caption{Nested NER results (F1) for ACE-2004, ACE-2005, GENIA and CNEC 1.0
    (Czech) corpora. {\bfseries Bold} indicates the best result, {\itshape
    italics} results above SoTA and \colorbox{gray!30}{gray background} indicates the main contribution. * uses different data split in ACE-2005. **
    non-neural model}
  \label{results:nested}
  \end{center}
  \end{small}
\end{table*}

\begin{table*}[!t]
  \begin{small}
  \begin{center}
  \catcode`@ = 13\def@{\bfseries}
  \catcode`! = 13\def!{\itshape}
  \begin{tabular}{l|c|c|c|c}
    model & English & German & Dutch & Spanish \\
    \hline
    \cite{Gillick2015}  & 86.50   & 76.22 & 82.84 & 82.95 \\
    \cite{Lample2016}   & 90.94   & 78.76 & 81.74 & 85.75 \\
    ELMo \cite{Peters2018}  & 92.22 & --  & --    & --    \\
    Flair \cite{Akbik}  & 93.09  & @88.32 & --   & --    \\
    BERT \cite{BERT}    & 92.80   & --    & --    & --    \\
    \hline
    LSTM-CRF             & 90.72 & 79.89 & !87.42 & !86.34 \\
    LSTM-CRF+ELMo        & 92.58 & --    & --     & --     \\
    LSTM-CRF+BERT        & 92.94 & 84.53 & !92.48 & !88.77 \\
    LSTM-CRF+Flair       & 92.25 & 82.35 & !88.31 & --     \\
    LSTM-CRF+BERT+ELMo   & 92.93 & --    & --     & --  \\
    LSTM-CRF+BERT+Flair  & !93.22 & 84.44 & @92.69 & --    \\
    LSTM-CRF+ELMo+BERT+Flair & @93.38 & --    & --     & --     \\
    \hline
    seq2seq              & 90.77 & 79.09 & !87.59 & !86.04 \\
    seq2seq+ELMo         & 92.43 & --    & --     & --     \\
    seq2seq+BERT         & 92.98 & 84.19 & !92.46 & @88.81 \\
    seq2seq+Flair        & 91.87 & 82.68 & !88.67 & --     \\
    seq2seq+BERT+ELMo    & 92.99 & --    & --     & --     \\
    seq2seq+BERT+Flair   & 93.00 & 85.10 & !92.34 & --     \\
    seq2seq+ELMo+BERT+Flair & 93.07 & --    & --     & --  \\
    \end{tabular}
    \caption{Flat NER results (F1) for CoNLL-2002 and CoNLL-2003. {\bfseries
    Bold} indicates best result, {\itshape italics} results above SoTA.}
    \label{results:flat}
  \end{center}
  \end{small}
\end{table*}

\section{Results}
\label{section:results}

Table~\ref{results:nested} shows the F1 score for the nested NER and
Table~\ref{results:flat} shows the F1 score for the flat NER.

When comparing the results for the nested NER in the baseline models (without the
contextual word embeddings) to the previous results in literature, we see that
\textbf{LSTM-CRF} reaches comparable, but suboptimal results in three out of
four nested NE corpora, while \textbf{seq2seq} clearly outperforms all the
known methods by a wide margin. We hypothesize that \textbf{seq2seq}, although
more complex (the system must predict multiple labels per token, including the
special label \texttt{"<eow>"}), is more suitable for more complex corpora. The
gain is most visible in ACE-2004 and ACE-2005, which contain extremely long
named entities and the level of ``nestedness'' is greater than in the other
nested corpora. According to \citet{Wang2018}, $39\%$~of~train sentences contain
overlapping mentions in ACE-2004, as opposed to $22\%$~of~train sentences with
overlapping mentions in GENIA. With shorter and less overlapping entities, such
as in GENIA, and ultimately in flat corpora, the simplicity of
\textbf{LSTM-CRF} wins over \textbf{seq2seq}.

We also report a substantial increase in the F1 score when recently published
contextual embeddings (ELMo, BERT, Flair) are added as pretrained word embeddings on input
\cite{Peters2018,BERT,Akbik} in all languages and corpora, although in the case of
CoNLL-2003 German, our results stay behind those of \citet{Akbik}.

\section{Conclusions}

We presented two neural architectures for nested named entities and a simple
encoding algorithm to allow the modeling of multiple NE labels in an enhanced
BILOU scheme. The LSTM-CRF modeling of NE multilabels is better suited for
putatively less-nested and flat corpora, while the sequence-to-sequence
architecture captures more complex relationships between nested and complicated
named entities and surpasses the current state of the art in nested NER on four
nested NE corpora. We also report surpassing state-of-the-art results with the recently published contextual word embeddings on both nested and flat NE corpora.

\section*{Acknowledgements}

The work described herein has been supported by OP VVV VI LINDAT/CLARIN project
of the Ministry of Education, Youth and Sports of the Czech Republic (project
CZ.02.1.01/0.0/0.0/16\_013/0001781) and it has been supported and has been
using language resources developed by the LINDAT/CLARIN project of the Ministry
of Education, Youth and Sports of the Czech Republic (project LM2015071).

We would like to thank the reviewers for their insightful comments.  

\bibliographystyle{acl_natbib}
\bibliography{acl2019}

\end{document}